\documentclass{amia}
\usepackage{graphicx}
\usepackage[labelfont=bf]{caption}
\usepackage[superscript,nomove]{cite}
\usepackage{color}
\usepackage{bbm}
\usepackage{amsmath}
\usepackage{hyperref}
\usepackage{setspace}

\begin{document}

\title{Adversarial Sample Enhanced Domain Adaptation: A Case Study on Predictive Modeling with Electronic Health Records}

\author{Yiqin Yu$^{1}$, Pin-Yu Chen$^{2}$, Yuan Zhou$^{1}$, Jing Mei$^{1}$}

\institutes{
    $^1$IBM Research, Beijing, China;
    $^2$IBM Research, Yorktown Heights, NY, USA\\
}

\maketitle

\noindent{\bf Abstract}

\textit{With the successful adoption of machine learning on electronic health records (EHRs), numerous computational models have been deployed to address a variety of clinical problems. However, due to the heterogeneity of EHRs, models trained on different patient groups suffer from poor generalizability. How to mitigate domain shifts between the source patient group where the model is built upon and the target one where the model will be deployed becomes a critical issue. In this paper, we propose a data augmentation method to facilitate domain adaptation, which leverages knowledge from the source patient group when training model on the target one. Specifically, adversarially generated samples are used during domain adaptation to fill the generalization gap between the two patient groups. The proposed method is evaluated by a case study on different predictive modeling tasks on MIMIC-III EHR dataset. Results confirm the effectiveness of our method and the generality on different tasks.}

\section*{1 Introduction}
Benefiting from the accumulation of large scale electronic health records (EHRs), numerous computational models have been developed to facilitate clinical decision support processes with improved quality of health care and decreased costs. Data mining and machine learning technologies are widely used to build computational models to address a variety of clinical problems such as predictive modeling for risk prediction, disease progression modeling, and patient phenotyping \cite{yadav2018mining,shickel2017deep,ng2014paramo}. Although advanced machine learning methods, especially deep learning models, have gained notable performance improvements on the patient group where they are trained on, nevertheless, they still suffer from poor generalizability when deployed in new environments (e.g., downgraded AUROC or accuracy on the new dataset). This is mainly due to the heterogeneity of EHRs among different patient groups\cite{hripcsak2013next}. 

There are several aspects that will lead to the heterogeneity. First, the distributions of variables in EHR data which are used as inputs of the model as well as the modeling targets often differ among different patient groups \cite{zhang2019time}. For example, mid-life weight was more closely associated with mid-life blood pressure than weight at age 20 years \cite{sundstrom2020weight}. In patients hospitalized with myocardial infarction, women had higher mortality than men within the same age group \cite{canto2012association}. Second, the heterogeneity can also be subject to data biases \cite{gianfrancesco2018potential} such as the occurrences of missing data caused by the irregular ordering of lab tests and patients whose data is not collected at regular intervals. In other words, due to the heterogeneity of EHRs, the curated data of a specific patient group used to train the model is not fully representative of what the model will encounter when it is deployed.

Domain adaptation \cite{ben2010theory,kouw2018introduction} (DA) is one of the key methodologies of transfer learning to leverage knowledge from a different but related domain as additional information when training models on the target domain, where the input and label spaces remain unchanged while the data distributions change among two domains. In the context of predictive modeling on EHRs, a \textit{domain} usually corresponds to a patient group. Consequently, DA can enable modeling and improve machine learning performance on a target patient group by mitigating the domain shift between this patient group and another different yet related patient group. DA has been shown in plenty of applications that it can improve the generalizability of models, especially when there lacks labels (data sample annotations) in the target domain \cite{kouw2018introduction,zhuang2019comprehensive}.

While there are plenty of works applying domain adaptation methodology on EHRs, the main focus is on medical images \cite{hamed2019domain,gu2019progressive,zhang2020unsupervised,shen2020domain}, and only a few works study structured longitudinal EHR data. In recent years, several researchers use recurrent neural network based models to learn domain invariant temporal relationships between domains on multivariate time series EHR data \cite{purushotham2016variational,alves2018dynamic,zhang2019time}. Mainly, the domain adversarial neural network (DANN) architecture \cite{ganin2016domain} is incorporated to deal with the domain discrepancy via adversarial training between the label classifier and the domain discriminator. Even though these DANN oriented models achieve relatively high performance on unsupervised tasks, the intrinsic mechanism remains unclear. For example, the feature extractor in DANN is supposed to learn the domain invariant representation, but the selection of network architecture for the feature extractor is arbitrary and the learnt representation is hard to explain. 

In this paper, we propose a data augmentation method with adversarially generated samples to facilitate domain adaptation. Adversarial samples are formed by applying small but intentionally worst-case (automated and guided) perturbations to samples from the training dataset \cite{goodfellow2015explaining}. By involving adversarial samples into the training process (so called adversarial training), the trained model can achieve improved adversarial robustness of the model (capacity of defending against adversarial attacks) \cite{goodfellow2015explaining,madry2018towards,Ding2020MMA}. In spite of focusing on the adversarial robustness, several works study the effect of adversarial samples on model generalizability during domain adaptation. One work \cite{kouw2018introduction} proposes a method to augment the source domain dataset by iteratively appending adversarial samples during the training process and result in better generalizability to other unseen domains. Another work \cite{liu2019transferable} generates adversarial samples with DANN to fill in the gap between the source and the target domain, and adversarially trains the deep classifiers to make consistent predictions on adversarial samples. While these works provide strong evidence that adversarial samples can be a promising
data augmentation mechanism for deep neural network based domain adaptation on image data, there still lacks comprehensive studies and performance analysis on simple but more explainable algorithms such as logistic regression and more heterogeneous data such as EHRs.

Motivated by plentiful and practical domain adaptation requirements with EHRs, in this paper we introduce the proposed adversarial sample enhanced domain adaptation method with a case study on predictive modeling tasks with EHRs. We implement the method based on logistic regression (LR), which is the most dominant algorithm that has been used extensively in health and biomedical problems. Comparing to deep neural networks, LR is simple, easy to interpret and with less training time and computational cost yet acceptable performance. Furthermore, the proposed domain adaptation framework evaluated on LR in this paper can be easily applied to other empirical risk minimization based deep neural networks. For the case study, we utilize existing logistic regression benchmarks \cite{harutyunyan2019multitask} based on MIMIC-III EHR dataset \cite{johnson2016mimic} with three types of predictive modeling tasks: in-hospital mortality prediction, length-of-stay prediction and phenotype classification. A series of domain adaptation experiments are performed to compare among different patient groups with different variants of our method. 
Experimental results show that our method improves the generalizability of the trained models, and the method has good generality cross different predictive modeling tasks.

\section*{2 Dataset and Task Description}
We perform our case study on the MIMIC-III EHR dataset \cite{johnson2016mimic}. MIMIC-III is the largest publicly available database of de-identified EHR from ICU patients. It contains data that covers 38,597 distinct adult patients and 49,785 hospital admissions. Based on its large volume of data size and variety of data types, a recent work \cite{harutyunyan2019multitask} constructed four predictive modeling benchmarks. Among them, the following tasks are selected to evaluate our domain adaptation methods:
\begin{itemize}
\item \textbf{In-hospital mortality prediction.} Predicting in-hospital mortality based on the first 48 hours of an ICU stay. As a binary classification task, it is mainly evaluated by area under the receiver operating characteristic (AUROC). 
\item \textbf{Length-of-stay prediction.} Predicting remaining time spent in ICU at each hour of stay. It is formed as a multi-class classification problem with 10 classes (one for ICU stays shorter than a day, seven day-long buckets for each day of the first week, one for stays of over one week but less than two, and one for stays of over two weeks). Cohen’s linear weighted kappa score \cite{cohen1960coefficient} is used for evaluation.
\item \textbf{Phenotype classification.} Classifying which of the 25 care conditions are present in a given patient ICU stay. These conditions include both acute conditions such as acute cerebrovascular disease, acute myocardial infarction, and chronic conditions such as chronic kidney disease, diabetes mellitus without complication and essential hypertension. As a multi-label classification problem, it takes macro-averaged AUROC as the main metric.
\end{itemize}
All tasks share common feature space consisting of 17 clinical variables such as capillary refill rate, diastolic blood pressure and glucose. For each variable, six different sample statistic features (minimum, maximum, mean, standard deviation, skew and number of measurements) are computed on seven different sub-sequences of a given time series (the full time series, the first 10\% of time, first 25\% of time, first 50\% of time, last 50\% of time, last 25\% of time and last 10\% of time). Totally, there are $17 \times 7 \times 6 = 714$ features per time series. To test the domain shifts among different patient groups, we follow a most recent domain shifts evaluation work \cite{thiagarajan2019understanding} on MIMIC-III and divide the whole patient group into the following sub-groups according to the heterogeneity of EHR data: 1) \textbf{Age}. Two sub-groups are created: \textit{Older}, which comprises of patients who are 60 years old and above, and \textit{Younger}, which included patients who are younger than 60; 2) \textbf{Gender}. There are also two sub-groups: \textit{Female} and \textit{Male}.

\section*{3 Method}
We first define the notations used in our method. A \textit{domain} $\mathcal{D}=(\mathcal{X}, \mathcal{Y})$ denotes a \textit{patient group} which is represented by a number of patients whose observed features are in the input space $\mathcal{X}$ and the target outcomes in the output space $\mathcal{Y}$. $f(\theta)$ denotes a predictive model trained with logistic regression on $\mathcal{D}$ with parameter $\theta$. We define the \textit{source} domain as $D_S=(X_S, Y_S)$, where $X_S \in \mathbbm{R}^{m \times d}$ and $Y_S \in \mathbbm{R}^{m \times k}$, and the \textit{target} domain as
$D_T=(X_T, Y_T)$, where $X_T \in \mathbbm{R}^{n \times d}$ and $Y_T \in \mathbbm{R}^{n \times k}$. Here $m$ and $n$ is the number of samples in $D_S$ and $D_T$, respectively, and $d$ is the number of features with $k$ being the number of target outcomes. A \textit{sample} $(x,y) \in (\mathcal{X},\mathcal{Y})$ is an individual record processed by the model. Note that the output space $\mathcal{Y}$ for different modeling tasks will vary. In this paper, we deal with three types of tasks:
\begin{itemize}
\item \textbf{Binary classification.} There is only one target outcome in the output space $\mathcal{Y}$ with value of true or false. So here we set $k=1$ and $\mathcal{Y}=\{0, 1\}$. For example, in in-hospital mortality prediction, the target outcome is whether (true or false) the death occurs during hospitalization.
\item \textbf{Multi-class classification.} There is only one target outcome in the output space, but there can be more than two classes. Each class in the target outcome is exclusive, which means that each training instance can only be labelled to one class. So here $k=1$ and $\mathcal{Y}=\{0, \cdots, C-1\}$ where $C$ is the number of classes. The task of length-of-stay prediction is a multi-class classification when dividing the actual length of stay for each patient into 10 classes. Note that in practice, the multi-class classification problem can be decomposed into the binary classification problem, each of which has the output space $\mathcal{Y}=\{0, 1\}$.
\item \textbf{Multi-label classification.} Here more than one target outcome exist in the task with $k > 1$ (e.g., phenotype classification). For each outcome, it could be either a binary classification or a multi-class classification. For each patient, more than one outcome can be set to true.
\end{itemize}
\subsection*{3.1 Logistic Regression}
We first consider modeling tasks on the same domain, say the domain $D=(X, Y)$. As in the aforementioned discussion, both multi-class and multi-label classification can be decomposed into the binary classification problem. So without loss of generality we use $k=1$ throughout the method part. All the formulas can be easily generalized to multi-class and multi-label classification problems. Given a sample $(x,y)$ from $D$, the conditional distribution of $y$ given $x$ in logistic regression is modeled as
\begin{equation}
    \label{eqn:lr}
    p(y|x;\theta)=\left[g\left(\theta^{T}x \right)\right]^{y} \cdot\left[1-g\left(\theta^{T}x \right)\right]^{1-y}
\end{equation}
where $\theta$ is the parameter of the model $f(\cdot)$, and $g(\cdot)$ is the logistic function $g\left(\theta^{T}x\right)=\frac{1}{1+e^{-\theta^{T} x}}$. The loss function can be written as
\begin{equation}
    \label{eqn:lr_loss}
    l_D(\theta)=-\log \prod_{i=1}^{n} p\left(y_{i} | x_{i} ; \theta\right)=-\sum_{i=1}^{n}\left[y_{i} \log g\left(\theta^{T} x_{i}\right)+\left(1-y_{i}\right) \log \left(1-g\left(\theta^{T} x_{i}\right)\right)\right]
\end{equation}
During the training process, the parameters are estimated by minimizing the loss function:
\begin{equation}
    \label{eqn:lr_loss_min}
    \theta=\arg \min _{\theta} l_D(\theta)
\end{equation}
\subsection*{3.2 Domain Adaptation with Logistic Regression}
Given the source domain $D_S=(X_S, Y_S)$, the predictive model $f_S(\theta_S)$ trained on $D_S$ and the target domain $D_T=(X_T, Y_T)$, the domain adaptation framework aims to train the target model $f_T(\theta_T)$ by leveraging knowledge from $D_S$. According to different parts of $D_S$ where $f_T$ can leverage knowledge from, there are mainly two scenarios. In the first scenario, the source domain data $(X_S, Y_S)$ is known and is used to compare domain discrepancy between $(X_S, Y_S)$ and $(X_T, Y_T)$. The second one is to leverage knowledge from $f_S$, in case that $(X_S, Y_S)$ can not be accessed when training $f_T$ on $D_T$. In this paper, we will focus on the second situation as it occurs frequently in our considered scenarios (e.g., due to privacy of EHR, only the source model is accessible, but not the raw data).  

Following the previous work \cite{chelba2006adaptation}, we involve the parameter $\theta_S$ of the logistic regression model $f_S$ trained on $D_S$ into the training process of $f_T$. Suppose $\theta_S$ is obtained beforehand via the following training process:
\begin{equation}
    \label{eqn:lr_src}
    \theta_S=\arg \min _{\theta_S} l_{D_S}(\theta_S).
\end{equation}
Then, $\theta_T$ is optimized on $(X_T, Y_T)$ with the following loss function:
\begin{equation}
    \label{eqn:lr_tar}
    \theta_T=\arg \min _{\theta_T} l_{D_T}(\theta_T) + \lambda \|\theta_{T}-\theta_{S}\|^{2},
\end{equation}
where the second part is a modified $L_2$ regularizer to penalize the discrepancy between parameters of the source model $\theta_S$ and parameters of the target model $\theta_T$. $\lambda$ is used to trade off the proportion of knowledge that can be leveraged from the treatment of the model discrepancy.

\subsection*{3.3 Adversarial Sample Enhanced Domain Adaptation}
The idea of using adversarial samples as a method of data augmentation during domain adaptation is motivated by the following findings: 1) minimizing the model training error with adversarial samples can potentially bound the error on the target domain \cite{lee2018minimax}, and 2) training with adversarial samples can aid in finding domain-invariance and generalizable data features. \cite{zhang2019theoretically,Ding2020MMA,ilyas2019adversarial}.

As in the preceding equation \ref{eqn:lr_tar}, domain adaptation evaluates the discrepancy between two domains by comparing model parameters $\theta_S$ and $\theta_T$. However, due to the heterogeneity of different domains, the discrepancy always exists. Moreover, with the increasing of domain discrepancy, the training process will fail to adapt. In order to deal with this issue, we generate adversarial samples during domain adaptation and use them as part of the training set to make the adaptation more smoothly. In other words, the adversarial samples play two roles. First, it makes the trained model more robust and have better generalization ability; second, it plays as the intermediate distribution between the source and the target domain, so that the knowledge transfer will be more smooth. 

More concretely, we follow the iterative fast gradient sign method \cite{goodfellow2015explaining,kurakin2017adversarial} to generate adversarial samples by applying perturbations in the direction of the gradient of the loss function defined in Eq.~\ref{eqn:lr_src} and Eq.~\ref{eqn:lr_tar}, with the goal of increasing the training loss:
\begin{equation}
    \label{eqn:adv}
    x_{0}^{a d v}=x, \quad x_{Z}^{a d v}=\operatorname{Clip}_{x, \epsilon}\left\{x_{Z-1}^{a d v}+\frac{\epsilon}{Z}\operatorname{sign}\left(\nabla_{x} l\left(\theta,x_{Z-1}^{a d v}, y\right)\right)\right\},
\end{equation}
where $Z$ is the number of iteration, $\operatorname{Clip}_{x, \epsilon}\{\cdot\}$ is a function to perform feature-wise clipping of the generated adversarial sample to make it being in $L_{\infty}$ $\epsilon$-neighbourhood of the source sample $x$, and $\operatorname{sign}(\cdot)$ refers to being positive (+1) or negative (-1) of the gradient. 
Adversarial samples are generated for both the source domain and the target domain with Eq.~\ref{eqn:adv} and $\theta_S$ and $\theta_T$ from Eq.~\ref{eqn:lr_src} and Eq.~\ref{eqn:lr_tar}, respectively. Firstly, we train the LR model with both normal samples $x_S$ and adversarial generated samples $x_S^{a d v}$ on the source domain to get $\theta_S$ with Eq.~\ref{eqn:lr_adv_src}. Then, $\theta_S$ is used in the domain adaptation process with both normal samples $x_T$ and adversarial generated samples $x_T^{a d v}$ on the target domain to obtain $\theta_T$ with Eq.~\ref{eqn:lr_adv_tar}.
The modified loss functions for the source domain model and domain adaptation on the target domain are as follows:
\begin{equation}
    \label{eqn:lr_adv_src}
    \theta_S=\arg \min _{\theta_S} l_{D_S}(\theta_S,x_S,y_S) + \alpha l_{D_S}(\theta_S,x_S^{a d v},y_S)
\end{equation}
\begin{equation}
    \label{eqn:lr_adv_tar}
    \theta_T=\arg \min _{\theta_T} l_{D_T}(\theta_T,x_T,y_T) + \alpha l_{D_T}(\theta_T,x_T^{a d v},y_T) + \lambda \|\theta_{T}-\theta_{S}\|^{2}
\end{equation}
where $\alpha \in (0, 1]$ is a weighting coefficient to adjust contributions from the adversarial samples.

\section*{4 Results}
We evaluate the proposed domain adaptation method on three predictive modeling tasks with four domain adaptation tasks on MIMIC-III. The sizes of samples on different patient groups for each modeling task are shown in Table \ref{tab:cohort_size}. Note that in the length-of-stay prediction tasks each sample is the data record collected hourly. 
For domain adaptation, there are four tasks with different patient groups as the source domain and the target domain: \textit{Older}$\to$\textit{Younger}, \textit{Younger}$\to$\textit{Older}, \textit{Female}$\to$\textit{Male}, and \textit{Male}$\to$\textit{Female}. For each domain, the data is further divided into the training set with 85\% samples and the test set with remain 15\% samples. For example, for domain \textit{Female} in in-hospital mortality prediction, the total sample number is 9510, and the sizes of the training set and the test set are 8071 and 1439, respectively.

All models are trained based on logistic regression (Eq.~\ref{eqn:lr_loss_min}) with hyperparameters: learning rate$=0.01$, batch size$=100$ and number of epochs$=100$. In order to intervene into the empirical risk minimization process to perform adversarial training and domain adaptation with LR, we rewrite the algorithm with PyTorch \cite{pytorch2020} instead of using scikit-learn \cite{sklearn2020} as in the benchmark \cite{harutyunyan2019multitask}. To make the series of experiments simple, we also remove the $L_2$ regularizer which leads to a slight downgrade of performance in our experiment. 

\begin{table}[h]
\centering
\caption{Size of prediction instances for different modeling tasks.}
  \label{tab:cohort_size}
  \begin{tabular}{|l|l|l|l|l|l|}
  \hline
    \textbf{Modeling Tasks}  & \multicolumn{2}{|l|}{\textbf{Gender}}  & \multicolumn{2}{|l|}{\textbf{Age}}  &  \textbf{Total}\\ \hline
      & \textit{Female}  & \textit{Male}  & \textit{Younger}  & \textit{Older}    &\\ \hline
    In-hospital mortality prediction  & 9510  &  11629  &  7387  &  13752  &  21139    \\ \hline
    Length-of-stay prediction  & 1520754  &  1930592 & 1279008  &  2172338 & 3451346   \\ \hline
    Phenotype classification  & 18477  & 23425  & 15929  & 25973  &  41902  \\ \hline
  \end{tabular}
\end{table}

\begin{figure}[h!]
  \centering
  \includegraphics[scale=0.46]{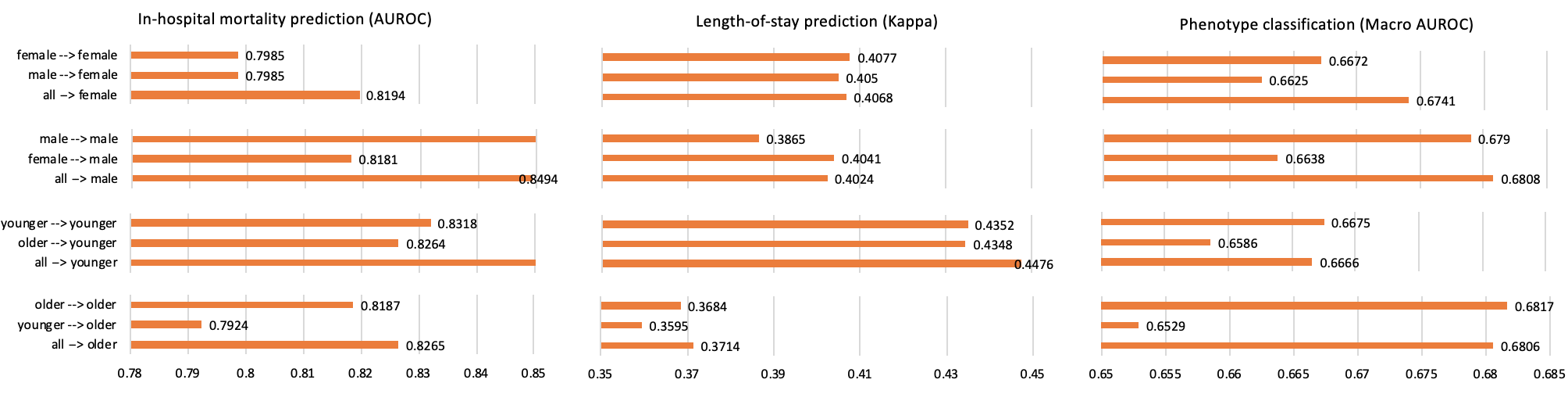}
  \caption{Domain shifts evaluated on different target domains.}
  \label{fig:domain_shift}
\end{figure}

\subsection*{4.1 Domain Shift Evaluation}
In order to better understand how the domain adaptation framework works, we first perform domain shift evaluation for all three modeling tasks with results shown in Figure~\ref{fig:domain_shift}. 
Each modeling task is evaluated with different measurements where AUROC is used for in-hospital mortality prediction, Kappa score is used for length-of-stay prediction, and Macro AUROC is used for phenotype classification. 
Taking the second bar ``male $\to$ female'' in in-hospital mortality prediction as an example, we first train the model with regular logistic regression (Eq. \ref{eqn:lr_loss_min}) on the domain \textit{Male} (trained on its training set and tested on its test set), and then test it with the test set of domain \textit{Female} to get the AUROC value as 0.799. 
For each modeling task, domain shift evaluation for the same target domain is grouped together. For instance, for the target domain \textit{Female}, there are three models being compared: the model trained on the same domain \textit{Female}, the model trained on the corresponding domain \textit{Male}, and the one trained on the full domain \textit{All} with data from \textit{Female} and \textit{Male} mixed together.

We can see that the model trained on \textit{All} and tested on \textit{Female} gets the best performance comparing to the model trained on sub-groups. It is consistent with common consensus that the size of training data has dominated impact on the performance. The most interesting finding is that the model trained on \textit{Male} is even slightly better than the one trained on \textit{Female}, which implies the greater possibility to transfer knowledge from \textit{Male} to \textit{Female}. Contrarily, with \textit{Male} as the target domain, the model trained on its own training set performs best among all models, which indicates \textit{Male} might have strong representability of the feature space for in-hospital mortality prediction. As a result, we can suspect that knowledge from \textit{Female} might have trivial effect on \textit{Male} during domain adaptation.

\begin{figure}[h!]
  \centering
  \includegraphics[scale=0.45]{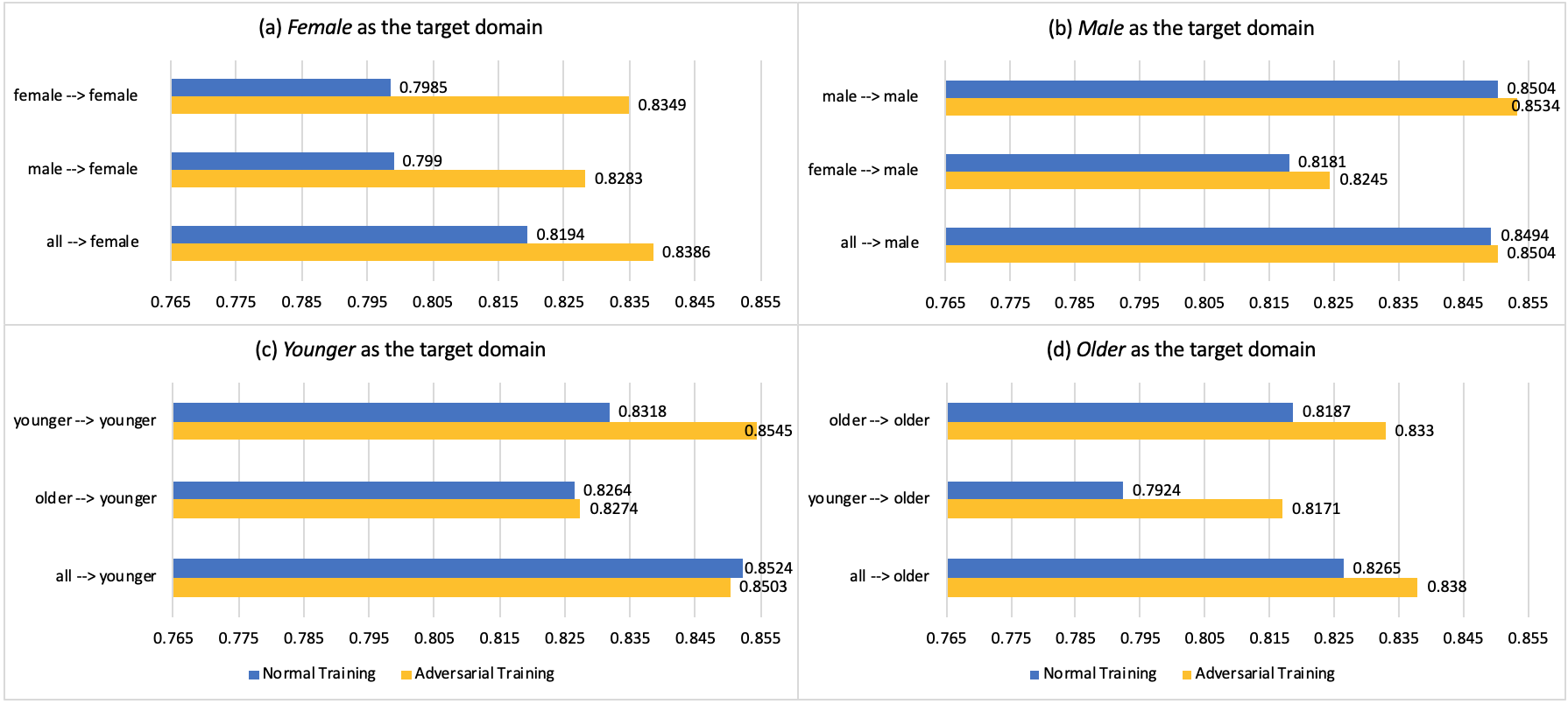}
  \caption{Adversarial training versus normal training for in-hospital mortality prediction (evaluated on AUROC).}
  \label{fig:nt_at}
\end{figure}
\begin{figure}[h!]
  \centering
  \includegraphics[scale=0.5]{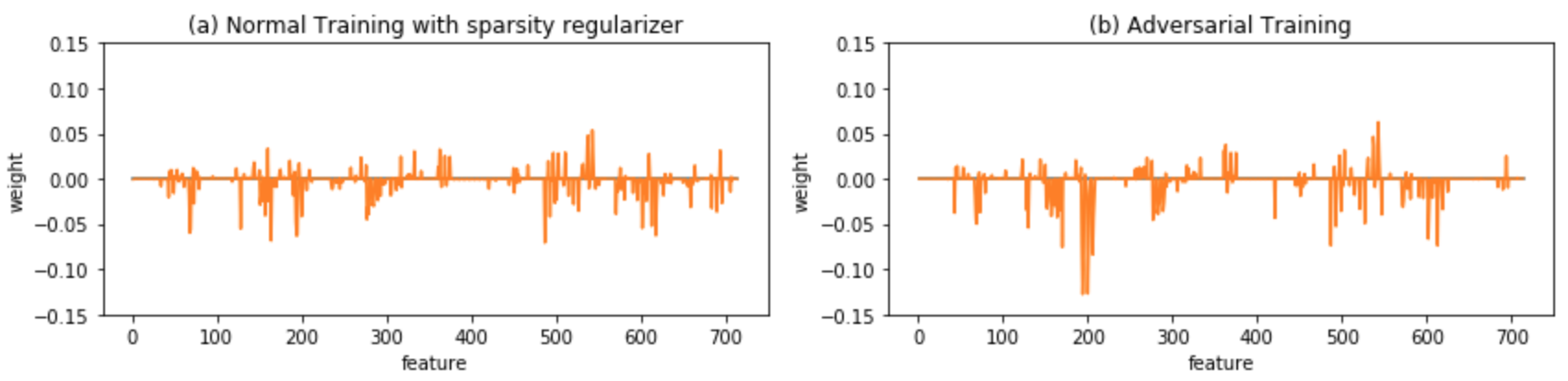}
  \caption{Comparison between $\theta$ from sparsity regularized normal training and $\theta$ from adversarial training for in-hospital mortality prediction on \textit{Female}$\to$\textit{Female} (cosine similarity=0.81).}
  \label{fig:weight_comp}
\end{figure}

\subsection*{4.2 Adversarial Sample as Data Augmentation}
Before using adversarial samples for domain adaptation, we also evaluate its effect on models trained on the source domain (Adversarial Training with Eq.~\ref{eqn:adv} and Eq.~\ref{eqn:lr_adv_src}) and tested on the target domain, and compare with the normal training methods (Normal Training with Eq.~\ref{eqn:lr_loss_min}). The results for in-hospital mortality prediction are shown in Figure~\ref{fig:nt_at} with weight comparison on \textit{Female}$\to$\textit{Female} in Figure~\ref{fig:weight_comp}. 
The adversarial training is performed with hyperparameters $\epsilon=0.1$, $T=20$ and $L_{\infty}$-norm as the perturbation type. Considering that the effect of adversarial samples on each patient group will differ, we test the adversarial training with different values of $\alpha$ among $(0, 1]$ and report the best performance in Figure~\ref{fig:nt_at}.

We can see that on almost all the tasks, the adversarial training models for in-hospital mortality prediction outperform the normal trained models except the case of \textit{All} $\to$ \textit{Younger}. Note that each adversarial training model achieves its best performance on different $\alpha$. For example, the AUROC for \textit{Male} $\to$ \textit{Male} achieves the largest value 0.8534 with $\alpha=0.1$ and monotonically decreases with larger $\alpha$. Unlike \textit{Male} $\to$ \textit{Male}, the AUROC for \textit{Male} $\to$ \textit{Female} get its highest value 0.8283 when $\alpha=0.3$, indicating that the adversarial samples of \textit{Male} play more important role on domain shift mitigation between \textit{Male} and \textit{Female}.
In Figure~\ref{fig:nt_at}, we can also see that the performance gain via adversarial training are greater in Figure~\ref{fig:nt_at} (a) than in Figure~\ref{fig:nt_at} (b). Recall the domain shifts evaluated for \textit{Male} in Figure~\ref{fig:domain_shift}, it might be due to the strong representability of the feature space of \textit{Male} for in-hospital mortality prediction. 
Another observation is that the adversarial training performs better on sub-groups than on the full groups (The only downgrading of adversarial training performance happens on \textit{All} $\to$ \textit{Younger}). It is reasonable because on sub-groups, the performance of the model is more likely to be limited by the training size, while on the full group, training data is relatively sufficient, so the effect of adversarial samples as a data augmentation method will be diminished. 

\subsection*{4.3 Adversarial Sample Enhanced Domain Adaptation}
\begin{figure}[h!]
  \centering
  \includegraphics[scale=0.46]{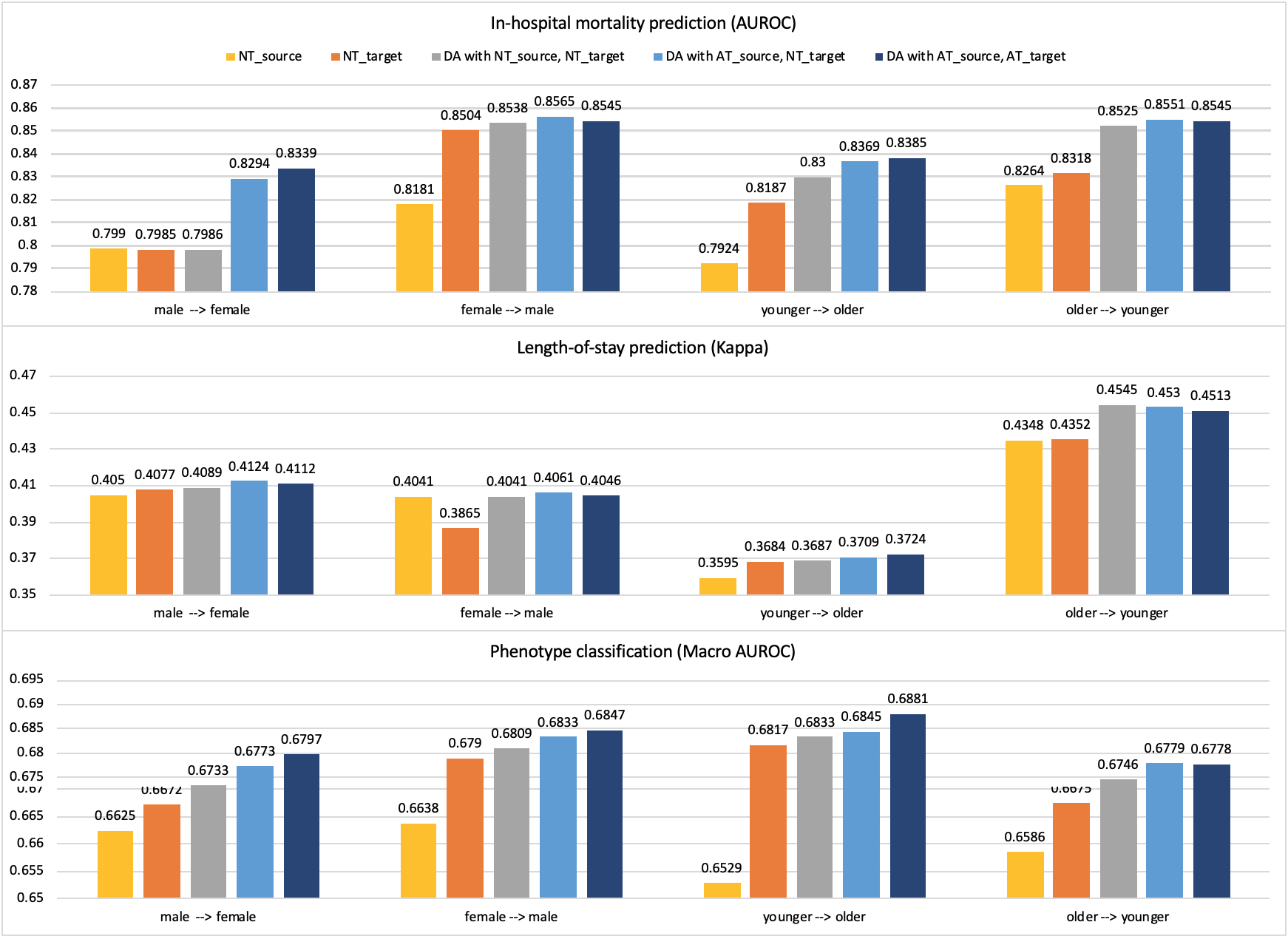}
  \caption{Domain adaptation results.}
  \label{fig:da}
\end{figure}

We implement our adversarial sample enhanced domain adaptation framework on all three predictive modeling tasks. More concretely, we first do adversarial training on the source domain, and then perform adversarial training with domain adaptation on the target domain. Finally the performance is evaluated on the test set of the corresponding target domain.
In Figure~\ref{fig:da}, we compare the performance of each domain adaptation task (e.g., \textit{Male}$\to$\textit{Female}) on two baseline settings and three domain adaptation variants:
\begin{itemize}
\setlength{\itemsep}{0pt}
\setlength{\parsep}{0pt}
\setlength{\parskip}{0pt}
\item NT\_source. Normal training on the source domain and testing on the target domain.
\item NT\_target. Normal training and testing on the target domain (the source domain is not used).
\item DA with NT\_source, NT\_target. First training on the source domain and then training with domain adaptation on the target domain without adversarial samples (Eq.~\ref{eqn:lr_tar}).
\item DA with AT\_source, NT\_target. Training procedure the same with the previous one, but use adversarial training on the source domain (Eq.~\ref{eqn:lr_adv_src}).
\item DA with AT\_source, AT\_target. Training procedure remaining with adversarial samples injected during domain adaptation on the target domain (Eq.~\ref{eqn:lr_adv_tar}). 
\end{itemize}
We can see that in almost all the circumstances, the three domain adaptation methods outperform the two baselines. It means that domain adaptation is capable to leverage knowledge from both the source domain and the target domain. 
Comparing ``DA with AT\_source, NT\_target'' with ``DA with NT\_source, NT\_target'', the adversarial training on the source domain works for domain adaptation in most tasks (11 out of 12 tasks). Note that the adversarial training on the source domain is not guided to optimize the domain adaptation task on the target domain. It plays only the role as data augmentation for the source domain. 
Between the later two variants of domain adaptation methods, we can also observe that the adversarial training on the target domain is able to improve the performance (9/12 tasks with adversarial training on the target domain have increased performance when compared to ``DA with AT source, NT target”). 

\section*{5 Discussion}
\textbf{Heterogeneity of EHRs.} The main reason of poor generalizability of predictive models trained on EHR data is the heterogeneity of EHRs. In this paper, we evaluate domain shifts mainly on the heterogeneity between patient groups with different age and gender groups. These two metrics are also frequently used in contemporary predictive modeling tasks to do risk stratification \cite{hieken2020sex,huynh2020age}. There are also other kinds of heterogeneity such as differences among label distributions and measurements discrepancies \cite{thiagarajan2019understanding}. We consider domain adaptation analysis on these scenarios as our future works. 

\textbf{Effect of adversarial samples.} Adversarial samples are originally proposed as an attack methods to test the adversarial robustness of a model, and then injected into the training data during the training process to defend the corresponding attacks \cite{goodfellow2015explaining,kurakin2017adversarial}. While the adversarial training ensures the adversarial robustness of the model, recent works \cite{su2018robustness,tsipras2019robustness,zhang2019theoretically} argue that there might be a trade-off between the adversarial robustness and the generalizability, reflected by higher robust accuracy and lower standard accuracy comparing with normal trained models. As these works are mainly evaluated on large volume image datasets, the effect of adversarial samples on more complex data types and feature distributions has not been well studied. For example, one work \cite{lei2019discrete} shows that adversarial training can improve the generalizability on text classification tasks. Our results also show that adversarial samples can help the generalizability especially for modeling tasks on sub-groups, implying its capacity as data augmentation on data with complex data types and feature distributions. Especially for logistic regression, it can be proved that jointly train the LR model with adversarial samples is equivalent to a sparsity regularized version of LR \cite{ar2020} (One can see in Figure~\ref{fig:weight_comp} that the $\theta$ of sparsity regularized LR and the one from adversarial trained LR are very similar with cosine similarity being 0.81). This finding also opens up a perspective for our further research.

\textbf{Domain adaptation types.} In this paper, we consider domain adaptation by transferring knowledge from the model trained on the source domain given the source domain data can not be accessed when training model on the target domain. 
There are a lot of other domain adaptation scenarios given whether certain elements in a domain $\mathcal{D}=\{(\mathcal{X}, \mathcal{Y}), f(\theta)\}$ are absent or not. For example, unsupervised domain adaptation deal with unlabeled target domain data (where $Y_T$ is absent) by learning a domain invariant joint representation on feature distributions of both the source domain and the target domain \cite{purushotham2016variational,alves2018dynamic,zhang2019time}. There are also strong interests in domain adaptation when the size of target domain data is very small, leading to subdivided research topics such as few-shot domain adaptation \cite{}. Limited to the length of the paper, we do not include the evaluation of our method on these domain adaptation tasks, but as our method achieves better performance on domains on sub-groups with smaller training size, it is expected to perform well under more restricted situations as well.

\textbf{Effect of domain adaptation.} Domain adaptation is proposed to leverage knowledge from other different but related domain(s) during model training on the target domain. One critical issue is to balance knowledge from the source domain and knowledge from the target training data. In this paper, we use grid search to optimize the hyperparameter $\lambda$ in Eq.~\ref{eqn:lr_adv_tar}. There are also some works aiming to automate the tuning of the hyperparameter, which will be brought into our future works. In this paper, we empirically prove that adversarial samples as a method of data augmentation can improve the generalizability for domain adaptation. It is worth to further analyse the relationship between this effect and other key attributes of the training task such as the data size as well as the heterogeneity of data.

\section*{6 Conclusion}
In this paper, we propose a domain adaptation framework and evaluate its effectiveness via a case study on different predictive modeling tasks with EHR data. Furthermore, adversarially generated samples are used as a data augmentation method in the domain adaptation procedure to improve its performance. Results show that our method improves the generalizability of predictive models on three predictive modeling tasks with the MIMIC-III EHR dataset.

\makeatletter
\renewcommand{\@biblabel}[1]{\hfill #1.}
\makeatother

\begin{spacing}{0.9}
\bibliographystyle{unsrt}
\bibliography{reference}
\end{spacing}

\end{document}